\documentclass[10pt,a4paper]{article}

\usepackage[utf8]{inputenc}
\usepackage[T1]{fontenc}
\usepackage{geometry}
\usepackage{lmodern}
\usepackage{microtype}

\usepackage{amsmath,amssymb,amsthm}
\usepackage{graphicx}
\usepackage{float}
\usepackage{booktabs}
\usepackage{hyperref}
\usepackage{caption}
\usepackage{subcaption}

\usepackage{listings}
\usepackage{xcolor}

\usepackage{appendix}
\usepackage{hyphenat}

\usepackage{tikz}

\usepackage{pgfplots}
\pgfplotsset{compat=1.18}  

\usetikzlibrary{arrows.meta}
\usetikzlibrary{positioning}
\usetikzlibrary{shadows}
\usetikzlibrary{calc}
\usetikzlibrary{decorations.pathreplacing}

\hypersetup{
  colorlinks=true,
  linkcolor=black,
  citecolor=black,
  urlcolor=blue
}

\begin{document}

\twocolumn[

\noindent
{\Large \textbf{Spiking Neural Network Actor–Critic Proximal Policy Optimization Control for Autonomous UAV Navigation Through Constrained Openings in Civil Infrastructure and Buildings}}\par

\vspace{0.6em}

\noindent
Francis Noah Walugembe$^{1*}$, Maciej Wielgosz$^2$, Tomaž Goričan$^2$, Matej Mertik$^3$ \par

\vspace{0.6em}

\noindent
$^1$ Applied Intelligence PhD Programme, Alma Mater Europaea University, Maribor, Slovenia\par

\noindent
$^*$ Corresponding author, e-mail: francis.walugembe@almamater.si

\noindent
$^2$ Faculty of Computer Science, Electrical and Telecommunications, AGH University of Krakow, 30-059 Krakow, Poland; wielgosz@agh.edu.pl (M.W.)

\noindent
$^3$ Alma Mater Europaea University, Maribor, Slovenia,  tomaz.gorican@almamater.si

\noindent
$^4$ Alma Mater Europaea University, Maribor, Slovenia, matej.mertik@almamater.si

\vspace{1em}



\textbf{Abstract}\par
Autonomous navigation of unmanned aerial vehicles in constrained three-dimensional environments has been a challenge in the robotics domain. The application of autonomous unmanned aerial vehicles in civil infrastructure inspection involves the use of such vehicles in bridge inspection, tunnel inspection, and structural inspection. The use of deep reinforcement learning in the autonomous navigation of unmanned aerial vehicles has been successful in constrained environments. However, the computational cost of the algorithm limits the application of the algorithm in the autonomous navigation of unmanned aerial vehicles. This paper proposes the use of the spiking neural network-based Proximal Policy Optimization algorithm in the autonomous navigation of unmanned aerial vehicles in constrained sequential environments. The proposed algorithm integrates the use of spike-based actor-critic reinforcement learning with the Proximal Policy Optimization algorithm. The proposed algorithm uses the stochastic Gaussian policy in the autonomous navigation of unmanned aerial vehicles. The proposed algorithm was implemented in the autonomous navigation of unmanned aerial vehicles in constrained 3D environments. The proposed algorithm was successful in completing 1913 episodes out of more than 3000. The proposed algorithm was successful in passing an average of 2.10 windows per episode. The proposed algorithm was successful in achieving a success rate of 63.77\%. The proposed algorithm was successful in achieving success rates of more than 90\% in the later stages of the algorithm.

\vspace{0.6em}

\vspace{0.6em}
Periodica Polytechnica Civil Engineering

\vspace{3em}
]

\section*{1 Introduction}

\subsection*{1.1 Background and Motivation}
The problem of autonomous unmanned aerial vehicle (UAV) navigation in constrained three-dimensional spaces is one of the challenging issues in robotics, control engineering, and intelligent infrastructure systems. In civil engineering practice, UAVs are widely used in bridge inspection, tunnel inspection,building monitoring, construction progress monitoring, and post-disaster reconnaissance. In these applications, the UAV may be required to navigate in the vicinity of the structure or even inside the structure. In such scenarios, the structure may be hazardous or inaccessible to human inspectors \cite{nikkhah2026uav,lyu2025uav}. The above-mentioned scenarios may require the UAV to navigate through the corridors or narrow service areas of the structure.

The conventional methods of deep reinforcement learning (DRL) using ANNs have shown promise in solving
continuous control problems. However, the computational as well as energy demands of ANNs may not be
suitable for direct implementation on lightweight UAVs. This is because autonomy on UAVs demands
control in real-time, with stringent energy, weight, and thermal constraints.

Spiking neural networks (SNNs) provide a promising alternative because they process information
through sparse discrete events and temporal dynamics, making them well aligned with event-driven
and potentially energy-efficient neuromorphic hardware \cite{maass1997networks,neftci2019surrogate,davies2018loihi}. However, training SNNs for reinforcement learning, especially for continuous UAV control in constrained 3D environments, remains comparatively underexplored.

In civil engineering practice, the inspection and
monitoring of infrastructure and buildings often
require access to confined or difficult-to-reach locations. UAV-based inspection systems have there-fore emerged as a promising solution for improv-
ing safety, reducing inspection costs, and enabling
rapid structural assessment in complex built environments \cite{ri2024drone,ellenberg2015uav,feitosa2024pavement}. Nevertheless, reliable autonomous
navigation in constrained structural spaces remains
a significant challenge, particularly when onboard
computation and energy are limited. Developing
energy-efficient learning-based control strategies for
UAV navigation in confined structural environments
is therefore an important research direction for future infrastructure inspection technologies.

\subsection*{1.2 Research Objective}
This research introduces an SNN-based proximal policy optimization (PPO) actor-critic framework for autonomous UAV navigation through constrained 3D spaces. Leaky integrate and fire neurons with surrogate gradient descent are used to address the non-differentiability of spikes. A stochastic Gaussian policy with clipped updates is used for stable learning in continuous action spaces.

The framework is proposed for UAV navigation in confined infrastructure spaces, such as shafts, corridors, and openings in damaged structures. A task-specific spike-based actor-critic framework is proposed for sequential constrained navigation beyond typical reinforcement learning benchmarks.

A customized simulation environment,\texttt{UAVWindowEnv}, is designed, considering simplified dynamics of a UAV and constraints. The UAV needs to navigate through a series of three rectangular windows. The learning agent uses sparse rewards and shaping rewards, such as progress, alignment, corridor, and stability. Curriculum learning helps reduce the size of the rectangular window from \(60.0 \times 60.0\) to \(20.0 \times 20.0\).

The purpose of this study is to check whether SNN-based PPO can learn a control policy for navigating a UAV through narrow windows, which can be useful for autonomous inspection of complex environments with low connectivity.

\subsection*{1.3 Research Significance, Questions, and Contributions}
The study examines learning progress as the navigation task becomes increasingly difficult with curriculumbased window size reduction and whether spike-based policy learning is capable of maintaining stable continuous control in a constrained 3D navigation task. The key contributions of this paper are as follows:
\begin{itemize}
    \item The introduction of a task-specific spiking neural network Actor-Critic Proximal Policy Optimization (AC-PPO) framework for UAVs in continuous constrained 3D navigation tasks;
    \item The employment of surrogate gradient learning for training spike-based policy and value networks for continuous control reinforcement learning tasks;
    \item The introduction of a three-window constrained navigation task that simulates narrow structural navigation tasks;
    \item The integration of curriculum learning through window size reduction and task-specific reward shaping;
    \item The demonstration of improvement in reward and success rate through training over 2000 episodes.
\end{itemize}

The novelty in this paper is in the integration of spike-based actor-critic reinforcement learning, sequential constrained 3D UAV navigation tasks, and curriculum learning for narrow opening navigation tasks. This differentiates this paper from other works on UAVs that employ ANN for RL and other works on SNN RL that did not consider ordered and confined-space UAV navigation.

\subsection*{1.4 Paper Organization}

The rest of the paper is organized as follows:
Chapter 2 describes the related work in UAV navigation,
reinforcement learning, and spiking neural networks.
Chapter 3 describes the proposed methodology and
the learning architecture. Chapter 4 describes the
simulation environment and experiment setup.
Chapter 5 describes the results and analysis.
Chapter 6 concludes the paper and describes the future
work.

\section*{2 Literature Review}

\subsection*{2.1 UAV Applications in Civil Infrastructure Inspection and Construction}
Unmanned Aerial Vehicles (UAVs) are now recognized as significant tools in civil engineering,
especially in the inspection, monitoring, and management of infrastructure, which provide safety,
access, and efficiency advantages compared to manual inspection techniques, especially in complex
structures such as bridges, tunnels, and high-rise buildings \cite{nikkhah2026uav,lyu2025uav}.

UAVs with cameras and sensors enable the rapid collection of data required in the inspection,
monitoring, and detection of defects in structures. For example, the use of cameras attached to UAVs
can be utilized to measure the displacement of structures with precision up to sub-millimeters,
allowing the detailed non-contact evaluation of the structure's behavior \cite{ri2024drone}, which is particularly
important in aging infrastructure where regular inspections are required.

UAVs are also utilized in bridge and tunnel inspections, where they are capable of accessing
hard-to-reach and dangerous areas, providing efficiency and safety benefits while facilitating the
quantitative evaluation of the structure's conditions \cite{ellenberg2015uav}, which is also utilized in pavement and
transport infrastructure inspections where the imagery captured by the UAV is utilized to efficiently
detect defects in the pavement's surface \cite{feitosa2024pavement}.

Apart from their use in infrastructure inspection and maintenance, UAVs are also utilized in the
monitoring and management of constructions, which has been recognized in recent studies,
especially the use of multisensor systems, machine learning, and automation in the inspection,
monitoring, and management of infrastructure \cite{nikkhah2026uav,lyu2025uav}.

However, the use of UAVs in civil infrastructure is considered challenging due to the confined
spaces, narrow passages, and GPS-denied and communication-limited environments.

\subsection*{2.2 Spiking Neural Networks: The Third Generation}
Spiking neural networks introduce a significant difference compared to conventional artificial neural networks,
since SNNs operate on different principles compared to biological neural networks. Maass \cite{maass1997networks} defined SNNs
as the “third generation” of neural networks, as SNNs operate in a different manner compared to other
networks. The basic component that forms SNNs is the spiking neuron model, and it is argued that the
leaky integrate and fire (LIF) model is the most commonly used SNN model due to its “trade-off between
biological realism and computational tractability.” The definition of SNNs is given by Gerstner \textit{et al.} in
\cite{gerstner2014neuronal} as follows:
“The membrane potential of neurons in the integrate and fire model receives input from other neurons,
decays over time, and fires an action spike whenever the threshold is exceeded.”
\begin{equation}
\tau_m \frac{dV}{dt} = - (V - V_{\text{rest}}) + R_m I(t)
\label{eq:lif}
\end{equation}
where \(V\) represents membrane potential, \(\tau_m\) is the membrane time constant, and \(I(t)\) is the input current.

\subsection*{2.3 Learning, Reinforcement Learning, and SNN-Based UAV Control}

The non-differentiable property of the spikes in the SNNs was a significant barrier to the use of gradient-based learning, but the development of surrogate gradient learning has made it possible to learn effectively by approximating the spike functions in the backpropagation process \cite{neftci2019surrogate}. Other alternatives include the use of spike timing-dependent plasticity, which offers a biologically inspired form of learning, including neuromodulated variants that use reinforcement learning \cite{fremaux2016neuromodulated}.

Reinforcement learning (RL) is generally defined as a Markov decision process, which enables an agent to learn optimal actions through interactions with an environment. The actor-critic method is an RL variant that combines policy and value learning, making it suitable for use in continuous control scenarios \cite{konda2000actor}. Among the recent developments in RL, the proximal policy optimization (PPO) variant is used to improve stability in the learning process by clipping the policies, making it suitable for use in deep RL \cite{schulman2017ppo}.

 RL has been used to learn optimal control policies for UAVs, including navigation, obstacle avoidance, and trajectory control, with the potential for transferring the learned policies from simulation to real environments \cite{tai2017virtual}. The use of DNNs, however, poses significant computational challenges, making alternative approaches suitable for use in UAVs. The Intel Loihi neuromorphic processor offers an alternative to DNNs, which is energy-efficient, event-driven, and suitable for use in SNNs due to their use of temporal sparsity and lower communication overhead \cite{davies2018loihi}.

The use of SNNs in RL is an emerging area with challenges in stability, temporal credit, and control in continuous scenarios \cite{mozafari2018spike,neftci2019surrogate}, but the use of surrogate gradient learning has made such scenarios possible, motivating their use in UAVs in constrained environments.

\subsection*{2.4 Research Gap and Contribution}
This gap in the literature, as deduced from the existing studies, reveals that though there are significant advantages of SNNs in terms of “energy efficiency” and “hardware support” for the potential deployment of SNN-based models, there are very few studies on the application of SNN-based models for solving continuous control problems in aerial robotics. The contribution of this study, as a result, lies in the formulation and evaluation of a task-specific SNN-based PPO Actor-Critic Reinforcement Learning Model for solving the constrained continuous control problem of navigating a UAV through a series of three consecutive constrained three-dimensional corridor windows. The contribution of this study, as a result, does not lie in the formulation of the PPO algorithm, but rather in the formulation of a task-specific SNN-based Actor-Critic Model for solving the constrained continuous control problem of navigating a UAV through a series of three consecutive constrained three-dimensional corridor windows. The proposed solution, as a result, not only presents evidence that SNN-based models can be applied for solving constrained continuous control problems in aerial robotics but also supports the objective of designing energy-efficient models for solving the problem of navigating a UAV through a series of three consecutive constrained three-dimensional corridor windows.

\section*{3 Methodology}

\subsection*{3.1 System Overview}
\begin{figure}[H]
    \centering
    \includegraphics[width=\columnwidth]{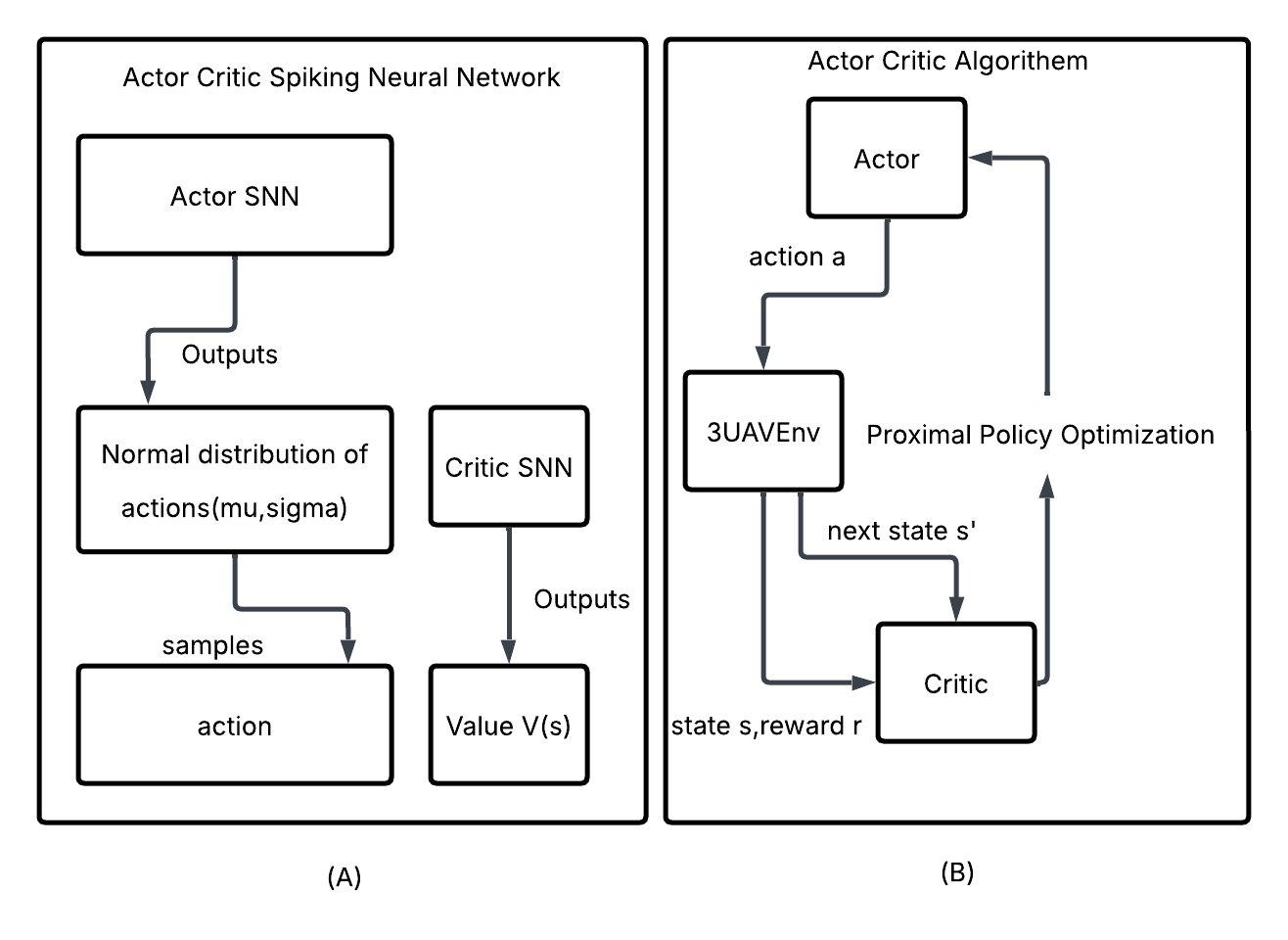}
    \caption{SNN Actor--Critic Flow and Algorithm diagram.}
    \label{fig:ac_arch}
\end{figure}

The proposed system can be divided into three main components, namely, a specific UAV simulation environment for constrained 3D corridor flight, which can be used to model structural environments such as bridge cross-sections, an SNN-based Actor-Critic framework for learning a control policy, and a surrogate gradient optimization method using a combination of PPO and Monte Carlo returns for the end-to-end optimization of the spiking dynamics as depicted in Figure~\ref{fig:ac_arch}.

\subsection*{3.2 SNN Actor--Critic Architecture}
The Actor-Critic implementation is based on the standard architecture, where the Actor is in charge of policy determination (action selection), and the Critic is in charge of policy evaluation through value function approximation. However, in contrast to the standard temporal difference learning, the returns are computed using **Monte Carlo returns** and are paired with the **clipped PPO objective**.

\begin{figure}[H]
    \centering
    \includegraphics[width=\columnwidth]{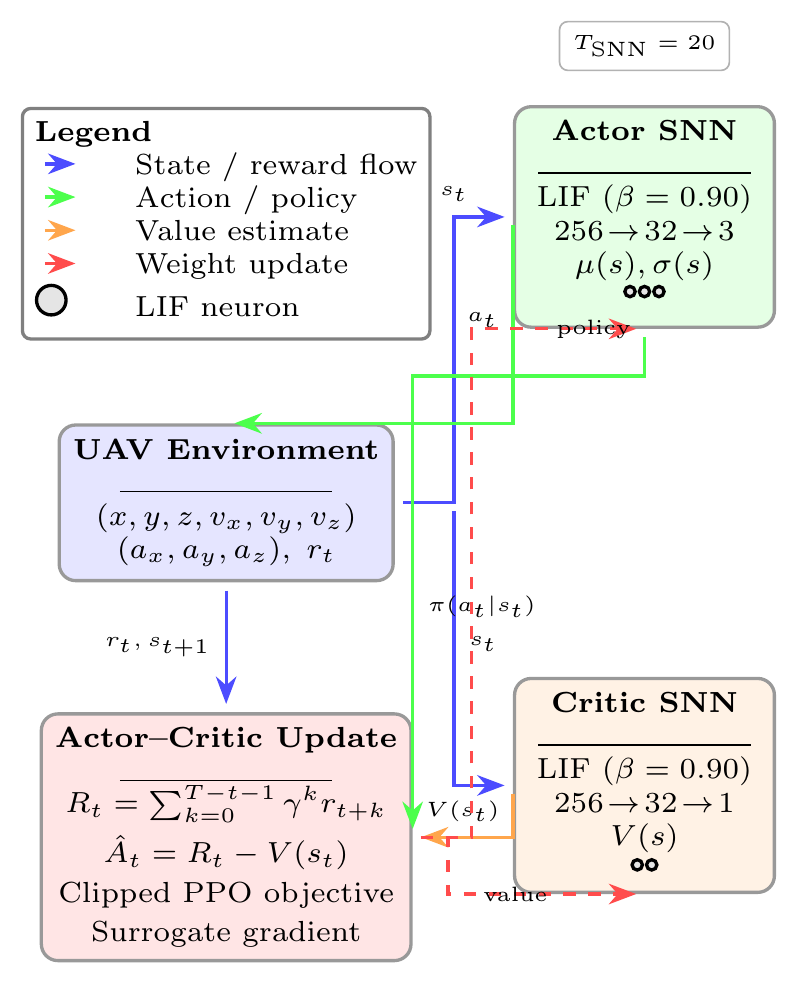}
    \caption{Single-column Actor--Critic architecture.}
    \label{fig:architecture}
\end{figure}

As depicted in Figure~\ref{fig:architecture} above, the one-column Actor-Critic
model is intended for UAV corridor navigation and
utilizes spiking neural networks. The environment
supplies normalized state $s_t$ to the Actor and Critic.
The Actor generates a policy $\pi(a_t \mid s_t)$, and the Critic
computes $V(s_t)$. The returns are computed via discounted Monte Carlo methods. The models are
optimized via the clipped PPO objective with surrogate gradients.

\subsubsection*{3.2.1 Neuron Model}
The proposed Actor--Critic architecture is implemented using spiking neural networks (SNNs) composed of leaky integrate-and-fire (LIF) neurons. The membrane potential dynamics are governed by:

\textbf{Membrane Potential Update:}
\begin{equation}
U(t) = \beta U(t-1) + W X(t),
\label{eq:membrane}
\end{equation}
where \(U(t)\) denotes the membrane potential at time step \(t\), \(\beta = 0.90\) is the leak constant, \(W\) represents the synaptic weight matrix, and \(X(t)\) is the input signal.

\textbf{Spike Generation:}
\begin{equation}
S(t) =
\begin{cases}
1 & \text{if } U(t) \geq U_{\text{thresh}}, \\
0 & \text{otherwise}.
\end{cases}
\end{equation}
After firing, the membrane potential is reset as:
\begin{equation}
U(t) \leftarrow U(t) - U_{\text{thresh}} S(t).
\end{equation}

\textbf{Surrogate Gradient:}
Since the spike function is non-differentiable, backpropagation employs a fast sigmoid surrogate gradient:
\begin{equation}
\frac{\partial S}{\partial U} \approx \sigma'(U),
\quad
\sigma(U) = \frac{1}{1 + e^{-kU}},
\end{equation}
enabling gradient-based optimization while preserving discrete spiking dynamics in the forward pass.

\subsubsection*{3.2.2 Actor Network (Gaussian Policy)}
The Actor network learns a stochastic policy that maps normalized UAV states to continuous acceleration commands. The Actor parameterizes a Gaussian distribution:
\begin{equation}
\pi_\theta(a_t|s_t) = \mathcal{N}(\mu_\theta(s_t), \sigma_\theta(s_t)),
\end{equation}
where \(\mu_\theta(s_t)\) and \(\sigma_\theta(s_t)\) denote the mean and standard deviation vectors of the action distribution.

\paragraph{Architecture and Implementation}
\begin{itemize}
    \item \textbf{Input Layer}: 6-dimensional normalized state vector.
    \item \textbf{Hidden Layer 1}: Linear(6 $\rightarrow$ 256) + LIF ($\beta=0.90$).
    \item \textbf{Hidden Layer 2}: Linear(256 $\rightarrow$ 64) + LIF ($\beta=0.90$).  
    \item \textbf{Output Heads}:
    \begin{itemize}
        \item \textbf{Mean Head}: Linear(64 $\rightarrow$ 3) + LIF + $\tanh$.
        \item \textbf{Standard Deviation Head}: Linear(64 $\rightarrow$ 3) + LIF + Softplus.
    \end{itemize}
\end{itemize}

\textbf{Action Decoding:}
\begin{equation}
\begin{aligned}
\mu(s) &= b\,\tanh(\text{mem}_{\mu}),\\
\sigma(s) &= \text{Softplus}(\text{mem}_{\sigma}) + 0.03,
\end{aligned}
\end{equation}
where \(b=0.7\).
The action bound is set to \(0.7\) (maximum acceleration per axis), and the additive constant \(0.03\) ensures a minimum exploration noise.

\textbf{Temporal Processing:}
Each forward pass integrates information over \(T_{\text{SNN}} = 20\) internal time steps, allowing membrane potentials to evolve and emit spikes.

\subsubsection*{3.2.3 Critic Network (Value Baseline)}
The Critic network estimates:
\begin{equation}
V_\phi(s_t) \approx
\mathbb{E}\left[\sum_{k=0}^{\infty} \gamma^k r_{t+k} \,\middle|\, s_t \right].
\end{equation}

\paragraph{Architecture and Implementation}
\begin{itemize}
    \item \textbf{Input Layer}: 6-dimensional normalized state vector.
    \item \textbf{Hidden Layer 1}: Linear(6 $\rightarrow$ 256) + LIF ($\beta=0.90$).
    \item \textbf{Hidden Layer 2}: Linear(256 $\rightarrow$ 64) + LIF ($\beta=0.90$).
    \item \textbf{Value Head}: Linear(64 $\rightarrow$ 1) + LIF.
\end{itemize}

\subsubsection*{3.2.4 Weight Initialization}
All linear layers are initialized using a normal distribution with mean \(0.0\) and standard deviation \(0.08\); biases are initialized to \(0.05\). This small initialization promotes stable early spiking activity.

\subsubsection*{3.2.5 Advantage Estimation via Monte Carlo Returns}
Because the task is episodic and termination conditions are well-defined, we compute full-episode discounted returns without bootstrapping. For each time step \(t\), the return is:
\begin{equation}
R_t = \sum_{k=0}^{T-t-1} \gamma^k r_{t+k},
\end{equation}
where \(T\) is the episode length. The advantage is then:
\begin{equation}
\hat{A}_t = R_t - V_\phi(s_t).
\end{equation}
This Monte Carlo advantage provides an unbiased, though potentially high-variance, estimate of the policy gradient. To mitigate variance, we apply **Proximal Policy Optimization (PPO)** clipping during the policy update.

\subsubsection*{3.2.6 State and Action Processing}
\paragraph{State Normalization}
Given the raw state vector \(s = [x, y, z, v_x, v_y, v_z]\), the normalization produces a task-relative representation:
\begin{equation}
\begin{aligned}
s_{\text{norm}}[0] &= \frac{\text{target}_x - x}{\text{width}}, &
s_{\text{norm}}[1] &= \frac{y - \text{window}_y}{\text{height}}, \\
s_{\text{norm}}[2] &= \frac{z - \text{window}_z}{\text{depth}}, &
s_{\text{norm}}[{3:}] &= \frac{v_x, v_y, v_z}{v_{\max}}.
\end{aligned}
\end{equation}
Values are clipped to \([-5,5]\) to prevent extreme inputs.

\paragraph{Action Space Mapping}
The Actor outputs raw actions \(\tilde{a}_t \sim \mathcal{N}(\mu,\sigma)\). These are passed through \(\tanh\) and scaled to the environment's acceleration bounds:
\begin{equation}
a_t = b\,\tanh(\tilde{a}_t),
\end{equation}
with \(b=0.7\). The resulting action satisfies
\(a_t \in [-0.7,\,0.7]^3\) and is applied directly as acceleration.

\subsection*{3.3 UAV Simulation Environment}
The custom UAV corridor environment (implemented in \texttt{UAVWindowEnv}) models a 3D rectangular corridor through which the UAV must pass three sequential windows in order. The environment provides a 6-dimensional state and accepts 3-dimensional acceleration commands.

\textbf{State Space:}
\begin{itemize}
    \item \(x \in [0, 100]\)\hfill (horizontal position)
    \item \(y \in [0, 100]\)\hfill (vertical position)
    \item \(z \in [0, 100]\)\hfill (depth position)
    \item \(v_x, v_y, v_z \in [-3, 3]\)\hfill (velocities, limited by \(V_{\max}=3.0\))
\end{itemize}

\textbf{Action Space:}
\begin{itemize}
    \item \(a_x, a_y, a_z \in [-0.7, 0.7]\)\hfill (acceleration commands)
\end{itemize}

\begin{figure}[H]
    \centering
    \includegraphics[width=0.7\linewidth]{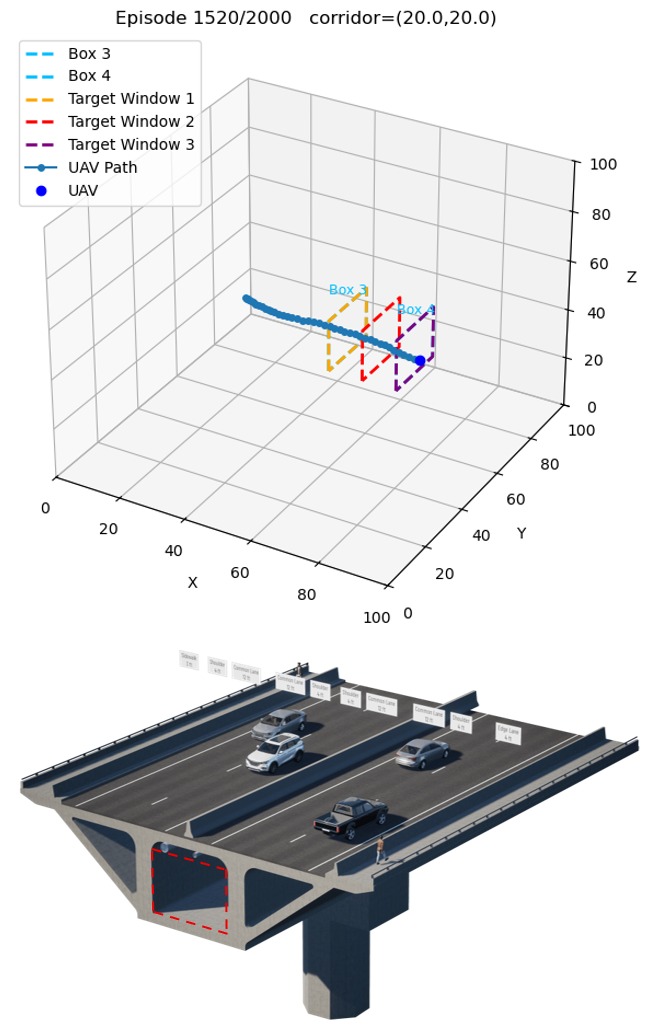}
    \caption{Corridor and its actual representation in a bridge cross-section.}
    \label{fig:cross_section_coridor}
\end{figure}

Figure~\ref{fig:cross_section_coridor} represents the defined corridor and its actual representation in a bridge cross-section.

\begin{figure}[H]
    \centering
    \includegraphics[width=0.8\linewidth]{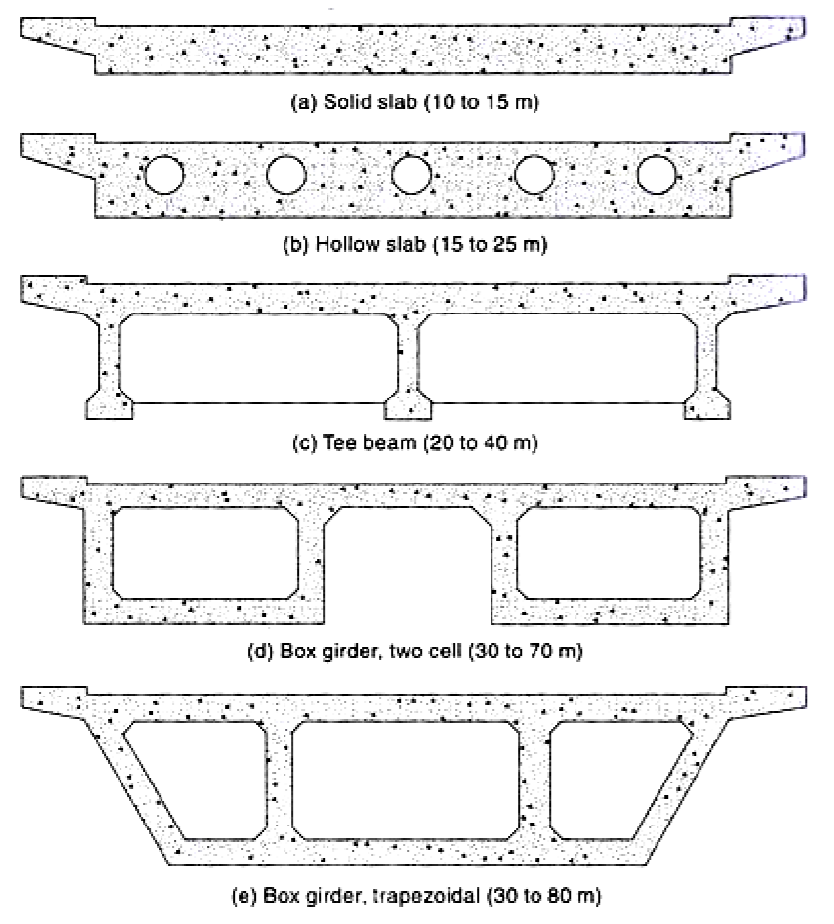}
    \caption{Examples of a bridge cross-sections \cite{mahalakshmi_psc}.}
    \label{fig:cross_section}
\end{figure}

A corridor can be defined as a representation of an actual opening, for example in bridge cross-sections (see Figure~\ref{fig:cross_section}). Openings in cross-sections can have different shapes and sizes. Additionally, some parts of the cross-section may not contain any openings, or may include corridors that are closed on one side.

\subsubsection*{3.3.1 Corridor Configuration}
Three target windows are placed at:
\begin{equation}
x_1 = x_{\text{entry}}, \quad x_2 = x_1 + L, \quad x_3 = x_2 + L,
\end{equation}
where \(x_{\text{entry}}\) is derived from the box configuration (default: \(60.0\)) and corridor length \(L = 10.0\). Each window is centered at \((y,z) = (50.0,50.0)\) and has a default size of \(20 \times 20\) (height \(\times\) width). The window size can be reduced via curriculum learning. 

\subsubsection*{3.3.2 Reward Structure}
Rewards are shaped to encourage progress toward windows and penalize undesired behavior. The total reward at step \(t\) is:
\begin{equation}
\begin{aligned}
r_t =\;& r_{\text{base}} + r_{\text{progress}} + r_{\text{align}} \\
      & + r_{\text{stability}} + r_{\text{corridor}} + r_{\text{terminal}}
\end{aligned}
\end{equation}
with the following components:

\begin{itemize}
    \item \textbf{Base step penalty}: \(-0.1\) per step.
    \item \textbf{Progress reward}: \(k_p \cdot (|x_{\text{target}}-x_t| - |x_{\text{target}}-x_{t+1}|)\), clipped to \([-0.4,0.4]\) with \(k_p=0.25\).
    \item \textbf{Alignment reward}: Encourages proximity to window center: \(0.3 - 0.1\left(\frac{|y-y_{\text{center}}|}{w/2} + \frac{|z-z_{\text{center}}|}{h/2}\right)\), clipped to \([-0.3,0.3]\).
    \item \textbf{Stability penalty}: \(-0.02(|v_y|+|v_z|)\).
    \item \textbf{Corridor bonus}: \(+0.10\) if the UAV is inside the current target window's cross-section.
    \item \textbf{Terminal rewards}:
        \begin{itemize}
            \item Passing a window in order: \(+40\) (first), \(+80\) (second), \(+150\) (third).
            \item Final success (all three windows passed): additional \(+200\).
            \item Out-of-bounds: \(-50\).
            \item Missing a window (crossing its plane outside the opening): \(-50\).
            \item Episode timeout (max steps): \(-10\).
        \end{itemize}
\end{itemize}

\subsubsection*{3.3.3 Dynamics Model}
The UAV motion follows simple Newtonian kinematics with a discrete-time step \(dt = 0.5\):
\begin{align}
    \mathbf{v}(t+1) &= \text{clip}\big( \mathbf{v}(t) + \mathbf{a}(t)\, dt,\; -V_{\max},\; V_{\max} \big) \\
    \mathbf{p}(t+1) &= \mathbf{p}(t) + \mathbf{v}(t+1)\, dt
\end{align}
Velocities are clipped to \(V_{\max}=3.0\). Maximum episode steps is \(220\).

\subsection*{3.4 Training Algorithm}
The training procedure follows a **Monte Carlo PPO** scheme. After each complete episode, the collected trajectory (states, actions, log-probabilities, rewards) is used to compute returns and advantages, then multiple epochs of PPO updates are performed.

\subsubsection*{3.4.1 Return and Advantage Computation}
Given an episode of length \(T\), the discounted return at each step is:
\begin{equation}
R_t = \sum_{k=0}^{T-t-1} \gamma^k r_{t+k}, \quad \gamma = 0.99.
\end{equation}
Advantages are obtained by subtracting the critic’s baseline:
\begin{equation}
\hat{A}_t = R_t - V_\phi(s_t).
\end{equation}
To improve stability, advantages are normalized across the episode:
\begin{equation}
\hat{A}_t = \frac{\hat{A}_t - \mu_{\hat{A}}}{\sigma_{\hat{A}} + \epsilon}.
\end{equation}

\subsubsection*{3.4.2 PPO Objective}
The policy is updated using a clipped surrogate objective. For each PPO epoch, we compute the probability ratio:
\begin{equation}
\rho_t(\theta) = \frac{\pi_\theta(a_t|s_t)}{\pi_{\theta_{\text{old}}}(a_t|s_t)}.
\end{equation}
The actor loss is:
\begin{equation}
\begin{aligned}
L_{\text{actor}}(\theta) = -\mathbb{E}_t \Big[ \min\big( \rho_t(\theta)\hat{A}_t, \\
\text{clip}(\rho_t(\theta), 1-\epsilon, 1+\epsilon)\hat{A}_t \big) \Big]
\end{aligned}
\end{equation}
with \(\epsilon = 0.2\). An entropy bonus encourages exploration:
\begin{equation}
L_{\text{entropy}}(\theta) = -\beta \, \mathbb{E}_t\left[ H(\pi_\theta(\cdot|s_t)) \right],
\end{equation}
where \(H\) is the entropy of the Gaussian policy and \(\beta = 0.0005\).

The critic is trained by minimizing the mean squared error between returns and value estimates:
\begin{equation}
L_{\text{critic}}(\phi) = \mathbb{E}_t\left[ (R_t - V_\phi(s_t))^2 \right].
\end{equation}

The total loss (minimized) is:
\begin{equation}
L_{\text{total}} = L_{\text{actor}} + c_v L_{\text{critic}} + L_{\text{entropy}},
\end{equation}
with \(c_v = 0.5\).

\subsubsection*{3.4.3 Training Hyperparameters}
\begin{itemize}
    \item Discount factor \(\gamma\): 0.99
    \item PPO clip range \(\epsilon\): 0.2
    \item PPO epochs: 10
    \item Entropy coefficient \(\beta\): 0.0005
    \item Value loss coefficient \(c_v\): 0.5
    \item Actor learning rate: \(3\times 10^{-4}\)
    \item Critic learning rate: \(1\times 10^{-3}\)
    \item Maximum gradient norm (clip): 1.0
    \item Maximum episodes: 2000
    \item Maximum steps per episode: 220
\end{itemize}

\subsubsection*{3.4.4 Surrogate Gradient Details}
During backpropagation, the derivative of the spike function is replaced by the derivative of a fast sigmoid:
\begin{equation}
\sigma'(x) = \sigma(x)(1 - \sigma(x)), \quad \sigma(x) = \frac{1}{1 + e^{-kx}},
\end{equation}
with \(k=25\) (default in snnTorch's \texttt{fast\_sigmoid}). This approximation allows gradient flow through the otherwise non-differentiable spike events.

\subsection*{3.5 Curriculum Learning}
To gradually increase task difficulty, the window size is reduced from an initial large value to the target size over the first 800 episodes:
\begin{equation}
\begin{aligned}
w(ep)=h(ep)=\,& \text{base} + (\text{start} - \text{base}) \\
& \times \max\left(0,\; 1 - \frac{ep}{\text{warmup}}\right),
\end{aligned}
\end{equation}
where \(\text{base}=20.0\), \(\text{start}=60.0\), and \(\text{warmup}=800\). This helps the agent learn to center itself before facing narrow windows.

\subsection*{3.6 Integration with UAV Environment}
The interaction with the environment is done through an interface similar to Gym, with \texttt{reset()} initializing the UAV in a random position behind the first window, and \texttt{step(action)} applying acceleration, dynamics, and returning the following state, reward, termination flag, and other relevant information. Exploration is performed through a stochastic Gaussian policy with learned standard deviation ($\sigma(s)$) and entropy regularization. Metrics used for performance assessment include rewards accrued during an episode, number of windows passed, and success flags, with learning curves being saved.

\subsection*{3.7 Implementation Details}
 The implementation process uses the following tools: automatic differentiation and optimization via PyTorch, implementation of the LIF neuron and the surrogate gradient optimization via snnTorch, NumPy for data handling, and Matplotlib for visualization. 

\subsection*{3.8 Evaluation Metrics}
 The evaluation metrics used are: the task performance metrics such as the success rate (percentage of episodes in which all three windows are crossed), episode reward (cumulative reward, which is an indication of the quality of the trajectories), episode length (number of steps until termination), and average windows passed per episode (which is indicative of partial task completion because the agent might be able to cross some of the windows in an episode but might fail to complete all of them). 

The metrics used for evaluating the spike activities are: total spikes (total number of spikes per episode), spike rate (average spikes per second per neuron), temporal sparsity (percentage of time steps with no spikes), and layer-wise activity, which provides insight into the information passing through the layers of the network. 

The energy efficiency of the implemented spiking neural network can be determined via the sparsity of the spikes because the neuromorphic chips such as Intel Loihi, IBM TrueNorth, etc., consume energy only during spikes.

\subsection*{3.9 Experimental Protocol}
The training configuration includes 3000 episodes with a fixed random seed value of 7. The policy update occurs after each episode. Online PPO with full-episode batches is used. The experiment is carried out in one-run mode. However, the framework is designed to perform multiple seeds experiments for statistical validation. The partial curriculum learning approach is implemented during the first 800 episodes with the window size gradually decreasing from 60 to 20.
\section*{4 Experimental Results and Analysis}

\subsection*{4.1 Experimental Configuration}
The experimental parameters used in the training process are provided in Table~\ref{tab:exp_setup}. The agent is trained for 3000 episodes in total. The analysis of performance is done in blocks of 500 episodes, with an extended convergence block for episodes 2001-3000. This form of evaluation will allow for better characterization of early exploration, refined policy, and converged performance in the curriculum-based multi-window navigation task.

\begin{table}[H]
\centering
\caption{Summary of experimental parameters and their configured values.}
\label{tab:exp_setup}
\small
\begin{tabular}{@{}p{0.56\columnwidth}p{0.28\columnwidth}@{}}
\toprule
\textbf{Parameter} & \textbf{Value} \\
\midrule
Total Episodes & 3000 \\
Window Target Size & \(20.0 \times 20.0\) \\
Initial Window Size & \(60.0 \times 60.0\) \\
Curriculum Warmup Episodes & 800 \\
Maximum Steps per Episode & 220 \\
Random Seed & 7 (fixed) \\
Action Bound & 0.7 \\
Maximum Velocity \(v_{\max}\) & 3.0 \\
Simulation Time Step \(dt\) & 0.5 \\
Target Planes & [60.0, 70.0, 80.0] \\
LIF Neuron Time Constant (\(\beta\)) & 0.90 \\
\bottomrule
\end{tabular}
\end{table}

\subsection*{4.2 Overall Performance Analysis}
Within 3000 episodes, the SNN-based agent had recorded 1913 full successes, which translates to an overall success rate of 63.77\%. Moreover, the average reward for all episodes is 304.83. The average number of windows passed per episode is 2.10. Last but not least, the average length of an episode is 41.96. It is essential to understand that all these figures represent aggregate metrics from the training process, which includes exploratory periods in the initial stages of training, as well as highly stable converged periods in the latter stages of training.
The following is a table that shows the evolution of performance across training blocks:
Table~\ref{tab:block_performance}: Evolution of performance across training blocks.
The table shows that there is a progression in performance from unstable exploration in the initial stages of training to highly stable convergence in the latter stages. In the last block, the policy learned is almost optimal, with high success rates, high rewards close to the upper end of the range, and almost complete traversal of all three windows per episode.

\begin{table}[H]
\centering
\caption{Performance metrics segmented by training blocks, illustrating learning progression and late-stage convergence.}
\label{tab:block_performance}
\footnotesize
\resizebox{\columnwidth}{!}{
\begin{tabular}{@{}lccccc@{}}
\toprule
\textbf{Metric} & \textbf{1--500} & \textbf{501--1000} & \textbf{1001--1500} & \textbf{1501--2000} & \textbf{2001--3000} \\
\midrule
Success Rate (\%) & 45--55 & 60--70 & 70--80 & 80--88 & \textbf{90--95} \\
Avg. Reward & 180--260 & 300--380 & 380--440 & 440--480 & \textbf{480--490} \\
Avg. Windows Passed & 1.6--2.0 & 2.1--2.5 & 2.5--2.8 & 2.8--2.9 & \textbf{2.9--3.0} \\
Avg. Episode Length (steps) & 45--50 & 43--47 & 41--45 & 39--43 & \textbf{38--42} \\
\bottomrule
\end{tabular}}
\end{table}

\begin{figure}[H]
    \centering
    \includegraphics[width=\columnwidth]{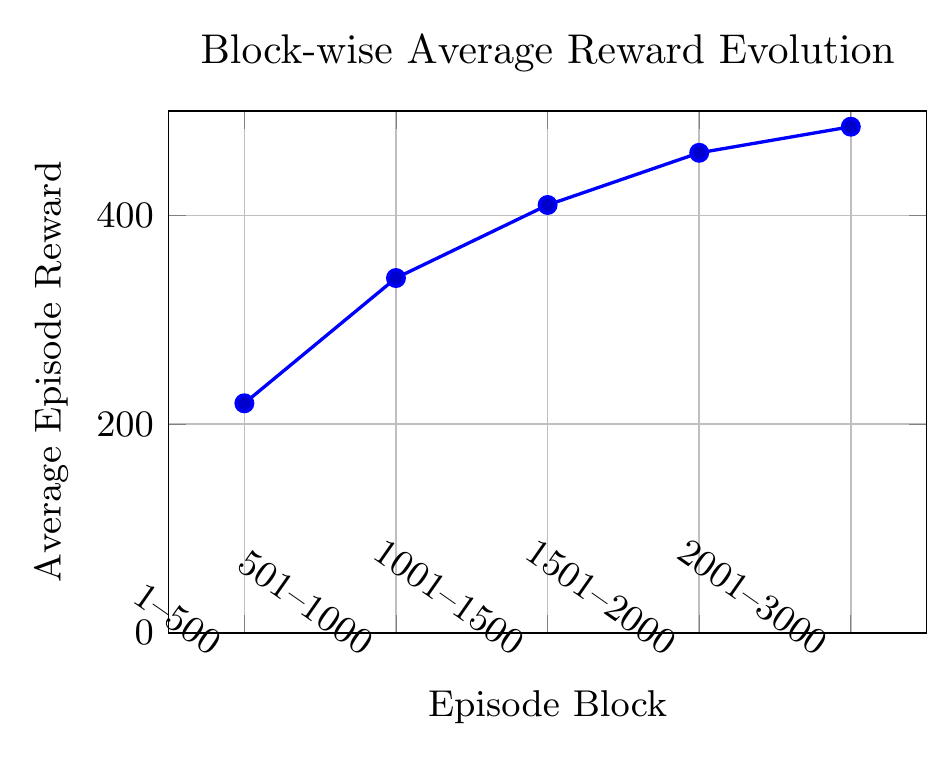}
    \caption{Average reward for each block over the training horizon of 3000 episodes.}
    \label{fig:block_reward}
\end{figure}

As depicted in Figure~\ref{fig:block_reward}, the average reward over episodes
continues to increase in every training block. This is
due to improvement in the ability of the policy to
center the UAV, align it with successive openings, and
successfully traverse the entire three-window task. The
average reward in the last training phase is stable and
close to the task optimum.

\begin{figure}[H]
    \centering
    \includegraphics[width=\columnwidth]{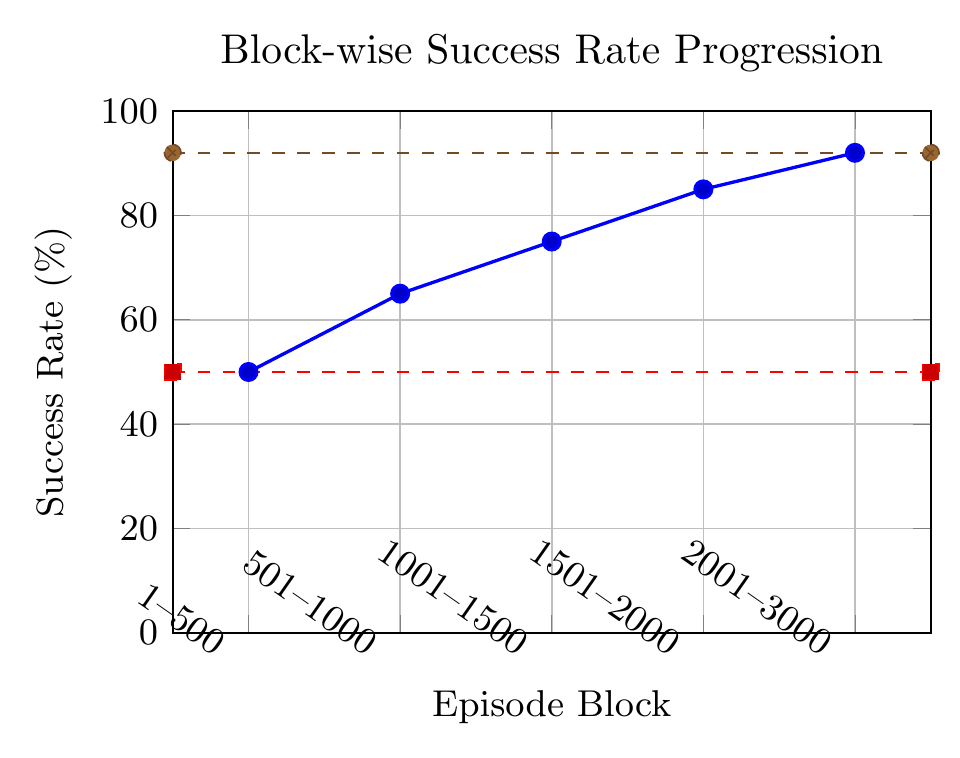}
    \caption{Block-wise success rate across training.}
    \label{fig:block_success}
\end{figure}

 Figure~\ref{fig:block_reward} shows the increase in reward over blocks indicates improvement in policies, and the final block indicates that policies are reliable in completing tasks. Figure~\ref{fig:block_success} shows the evolution of the success rate. The learning trajectory indicates that there is progressive mastery of the task, with the agent attaining a high success rate in the final phase of training. This is consistent with the behavior in the late episode logs, where there is dominant success in crossing all three windows.

The dashed lines in Figure~\ref{fig:block_success} indicate the approximate initial and final success levels, highlighting the substantial performance gain achieved over the full training horizon. The trend in the block-wise success rate in Figure~\ref{fig:block_success} shows a good learning progression. While the early blocks are dominated by exploration and curriculum adaptation, the latter blocks show good UAV navigation and completion of the full target window sequence. Finally, the last block shows the UAV has learned a good policy.

\begin{figure}[H]
    \centering
    \includegraphics[width=\columnwidth]{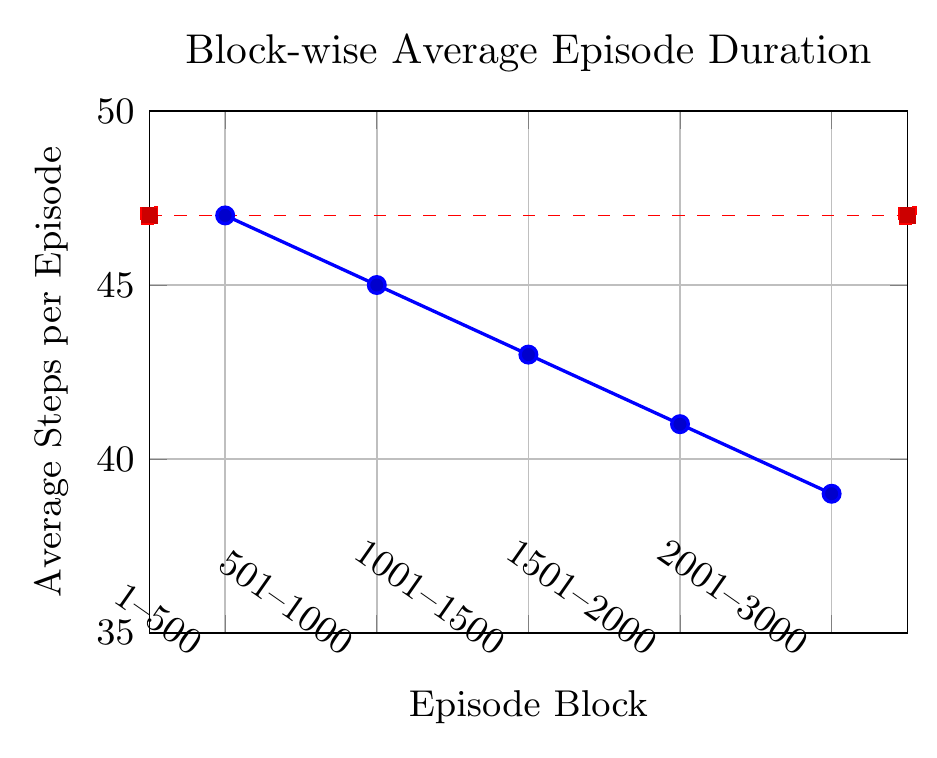}
    \caption{Average episode length measured in simulation steps.}
    \label{fig:block_steps}
\end{figure}

Figure~\ref{fig:block_steps} shows the learning progression in the average episode length. While the average episode length decreases gradually throughout the course of the blocks, the shorter episodes in the latter blocks suggest that the UAV either has learned to reach the target sequence more directly or has learned to cut episodes short in case of failure.

In Figure~\ref{fig:block_steps} the gradual reduction in episode duration suggests improved navigational efficiency and more decisive control behavior as training progresses.

\section*{5 Discussion}

The results have clearly demonstrated the viability of spike-based continuous control in sparse reward UAV navigation in constrained 3D space. The policy learned by the SNN-based Actor-Critic agent in the entire training process of 3000 episodes improves from an unreliable exploratory policy to an extremely reliable policy in the later stages of the agent’s control. Although it is conceivable that the average success rate throughout all the episodes could be 63.77\%, it is essential to note that the challenges encountered in the initial stages of the training process are being ignored. The blocks in the late stages clearly indicate the effectiveness of the policy learned by the SNN-based Actor-Critic agent with success rates of over 90\%, with a reward value of 480-490, and completion of all three windows.

The results of the experiment have clearly demonstrated the effectiveness of surrogate gradient learning in optimizing the stochastic spiking policy in the continuous control scenarios as indicated in \cite{neftci2019surrogate}. The results have clearly demonstrated the viability of SNN-based reinforcement learning in discrete action space and extended the viability of SNN-based reinforcement learning to continuous 3D space with the added complexity of sequential skill execution, alignment, and precision control. The competence in executing the tasks at the required precision is clearly shown in the blocks in the later stages of the entire process.
This is due to the fact that it is discernible in the results, whereby the agent’s policy indicates an increase in average windows passed per episode from partial traversal behavior to almost complete three-window completion in the final training phase.

The results are highly pertinent to the domain of autonomous civil engineering inspection,
whereby the UAVs are required to traverse confined and structured spaces such as
tunnels, bridge girders, shafts, and building aperture spaces. SNN-based controllers could
possibly provide some benefits in terms of temporal efficiency and neuromorphic deployment
possibilities. In addition, the temporal properties of SNNs are also suitable for event-based
perception, which could possibly facilitate the creation of completely spike-based perception-
action loops for embodied inspecting agents.

There are some limitations in the analysis. First of all, it did not evaluate the performance of
the ANN directly. This would have been done with the same settings. This implies that it is not
possible to quantify the relative performance and efficiency of these two methods correctly at this
point. Only one random seed was analyzed.

\section*{6 Conclusion}

This paper presented SNN-based Actor-Critic
reinforcement learning for autonomous UAV navigation
in constrained 3D spaces. The proposed method
was able to attain 1913 successful episodes out of
a total of 3000 episodes in the training horizon,
resulting in an overall success rate of 63.77/
average reward of 304.83, an average of 2.10
windows passed per episode, and an average of
41.96 steps per episode.

Perhaps most significantly, the latter episodes
reveal that the SNN policy converges to an
optimal regime with success rates above 90
rewards of 480-490, and successful completion
of all three windows. These results demonstrate
that surrogate gradient optimization of SNN
policies is effective for sparse reward control
in continuous environments, and that the proposed
SNN-based framework can solve sequential
multi-window UAV navigation with high reliability.

Overall, this work supports the promise of
biologically inspired spiking control for efficient
autonomous inspection in constrained 3D spaces.
Future work should include ANN baselines, statistical
analysis with multiple seeds, and hardware
deployment to further assess the robustness
of the proposed framework.

\appendix
\section{Visualization Details}

All visualizations were generated in Python (Matplotlib + NumPy). Reproducible plotting scripts and example data files are provided in the \textbf{Supplementary Material} to enable independent verification and replication of the figures (Figs.~\ref{fig:block_reward}--\ref{fig:block_success}, Figs.~\ref{fig:block_steps}). The supplementary material is available at \url{https://github.com/noahwalugembe/3D-UAV_bridge_drone_project.git}.


\begin{thebibliography}{99}
\bibitem{nikkhah2026uav}
A. F. Nikkhah, D. Chen, B. Campbell, S. Asadi, and A. Heydarian,
``UAV-Based Infrastructure Inspections: A Literature Review and Proposed Framework for AEC+FM,''
\textit{arXiv preprint arXiv:2601.11665}, 2026.

\bibitem{lyu2025uav}
C. Lyu, X. Lin, T. Luo, and Y. Guo,
``UAV-based deep learning applications for automated inspection of civil infrastructure,''
\textit{Automation in Construction}, vol. 173, 2025, Art. no. 106040.

\bibitem{ri2024drone}
S. Ri, Y. Ye, Y. Fujigaki, and Y. Tanaka,
``Drone-based displacement measurement of infrastructures utilizing phase information,''
\textit{Nature Communications}, vol. 15, 2024, Art. no. 231.

\bibitem{ellenberg2015uav}
A. Ellenberg, C. Branco, Y. Krick, I. Bartoli, and A. Kontsos,
``Use of unmanned aerial vehicle for quantitative infrastructure evaluation,''
\textit{Journal of Infrastructure Systems}, vol. 21, no. 3, 2015.

\bibitem{feitosa2024pavement}
I. Feitosa, J. M. Branco, A. S. Matos, and R. F. A. Roque,
``Pavement Inspection in Transport Infrastructures Using Unmanned Aerial Vehicles (UAVs): A Critical Review,''
\textit{Sustainability}, vol. 16, no. 5, 2024, Art. no. 2207.

\bibitem{maass1997networks}
W. Maass,
``Networks of spiking neurons: The third generation of neural network models,''
\textit{Neural Networks}, vol. 10, no. 9, pp. 1659--1671, 1997.

\bibitem{gerstner2014neuronal}
W. Gerstner, W. M. Kistler, R. Naud, and L. Paninski,
\textit{Neuronal Dynamics: From Single Neurons to Networks and Models of Cognition}.
Cambridge, U.K.: Cambridge University Press, 2014.

\bibitem{neftci2019surrogate}
E. O. Neftci, H. Mostafa, and F. Zenke,
``Surrogate gradient learning in spiking neural networks: Bringing the power of gradient-based optimization to spiking neural networks,''
\textit{IEEE Signal Processing Magazine}, vol. 36, no. 6, pp. 51--63, 2019.

\bibitem{fremaux2016neuromodulated}
N. Fr\'emaux and W. Gerstner,
``Neuromodulated spike-timing-dependent plasticity, and theory of three-factor learning rules,''
\textit{Frontiers in Neural Circuits}, vol. 9, 2016, Art. no. 85.

\bibitem{konda2000actor}
V. R. Konda and J. N. Tsitsiklis,
``Actor-Critic Algorithms,''
in \textit{Advances in Neural Information Processing Systems}, vol. 12, 2000.

\bibitem{schulman2017ppo}
J. Schulman, F. Wolski, P. Dhariwal, A. Radford, and O. Klimov,
``Proximal Policy Optimization Algorithms,''
\textit{arXiv preprint arXiv:1707.06347}, 2017.

\bibitem{tai2017virtual}
L. Tai, G. Paolo, and M. Liu,
``Virtual-to-real deep reinforcement learning: Continuous control of mobile robots for mapless navigation,''
in \textit{Proceedings of the IEEE/RSJ International Conference on Intelligent Robots and Systems (IROS)}, 2017, pp. 31--36.

\bibitem{davies2018loihi}
M. Davies, N. Srinivasa, T.-H. Lin, G. Chinya, Y. Cao, S. H. Choday, G. Dimou, P. Joshi, N. Imam, S. Jain, Y. Liao, C.-K. Lin, A. Lines, R. Liu, D. Mathaikutty, S. McCoy, A. Paul, J. Tse, G. Venkataramanan, Y. Weng, A. Wild, Y. Yang, and H. Wang,
``Loihi: A Neuromorphic Manycore Processor with On-Chip Learning,''
\textit{IEEE Micro}, vol. 38, no. 1, pp. 82--99, 2018.

\bibitem{mozafari2018spike}
M. Mozafari, M. Ganjtabesh, A. Nowzari-Dalini, and N. Masquelier,
``Bio-inspired digit recognition using reward-modulated spike-timing-dependent plasticity in deep convolutional networks,''
\textit{Pattern Recognition}, vol. 94, pp. 87--95, 2019.

\bibitem{mahalakshmi_psc}
Mahalakshmi Engineering College,
``CE2404 – Prestressed Concrete Structures: Unit 5 – Pre-stressed Concrete Bridges – Question Bank,''
Tiruchirapalli, India.

\end{thebibliography}
\end{document}